\documentclass[10pt,twocolumn,letterpaper]{article}
\usepackage[lined,boxed,commentsnumbered,ruled]{algorithm2e}
\usepackage{multirow}
\usepackage{bbding}
\usepackage{iccv}              % To produce the CAMERA-READY version
\definecolor{iccvblue}{rgb}{0.21,0.49,0.74}
\usepackage[pagebackref,breaklinks,colorlinks,allcolors=iccvblue]{hyperref}

\def\paperID{8040} % *** Enter the Paper ID here
\def\confName{ICCV}
\def\confYear{2025}

\title{Autoregressive Denoising Score Matching is a Good Video Anomaly Detector}

\author{\textbf{Hanwen Zhang\footnotemark[1] , Congqi Cao\footnotemark[1] \footnotemark[2] , Qinyi Lv, Lingtong Min, Yanning Zhang}\\
Northwestern Polytechnical University, Xi’an Shaanxi, 710129, China \\
{\tt\small zhwwww@mail.nwpu.edu.cn, 
\tt\small congqi.cao@mail.nwpu.edu.cn,
\tt\small \{lvqinyi, minlingtong, ynzhang\}@nwpu.edu.cn
}}

\begin{document}
\maketitle
\renewcommand\thefootnote{\fnsymbol{footnote}}
\footnotetext[1]{These authors contributed equally to this work.} %对应脚注[1]
\footnotetext[2]{Corresponding author.} %对应脚注[2]
\begin{abstract}
Video anomaly detection (VAD) is an important computer vision problem. Thanks to the mode coverage capabilities of generative models, the likelihood-based paradigm is catching growing interest, as it can model normal distribution and detect out-of-distribution anomalies. However, these likelihood-based methods are blind to the anomalies located in local modes near the learned distribution. To handle these ``unseen" anomalies, we dive into three gaps uniquely existing in VAD regarding scene, motion and appearance. Specifically, we first build a noise-conditioned score transformer for denoising score matching. Then, we introduce a scene-dependent and motion-aware score function by embedding the scene condition of input sequences into our model and assigning motion weights based on the difference between key frames of input sequences. Next, to solve the problem of blindness in principle, we integrate unaffected visual information via a novel autoregressive denoising score matching mechanism for inference. Through autoregressively injecting intensifying Gaussian noise into the denoised data and estimating the corresponding score function, we compare the denoised data with the original data to get a difference and aggregate it with the score function for an enhanced appearance perception and accumulate the abnormal context. With all three gaps considered, we can compute a more comprehensive anomaly indicator. Experiments on three popular VAD benchmarks demonstrate the state-of-the-art performance of our method. Code is available at \href{https://github.com/Bbeholder/ADSM}{https://github.com/Bbeholder/ADSM}.

\end{abstract}

% Meanwhile, we embed the scene condition of input sequences into our model to estimate a scene-dependent score function. Then, we assign motion weights to the score function based on the difference between key frames of input sequences. Next, we propose a novel autoregressive denoising score matching mechanism for inference, in which we autoregressively inject intensifying Gaussian noise into the denoised data and estimate the corresponding score function to form a more compact distribution. Finally, we compare the denoised data with the original data to get a difference and aggregate it with the score function for an enhanced appearance perception. With all three gaps considered, we can compute a more comprehensive anomaly indicator during inference. Experiments on three popular VAD benchmarks demonstrate the state-of-the-art performance and effectiveness of our method.    
\section{Introduction}
\label{sec:intro}

\begin{figure}
  \centering
  \includegraphics[width=0.98\columnwidth]{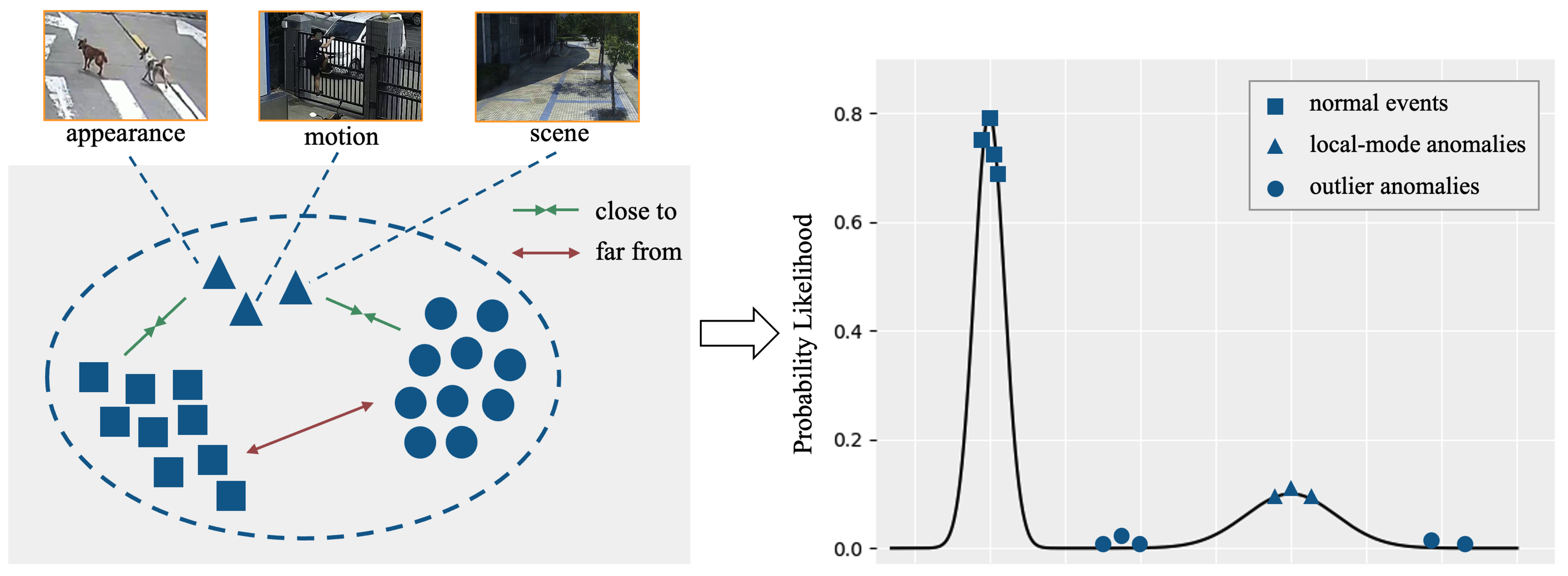}
  \caption{An illustration of the local modes of anomalies. We explore two regions in which the anomalies are residing. The likelihood-based method is highly sensitive to regions of anomalies with inherently low probability likelihood while blind to anomalies located in local modes near the learned distribution.}
  \label{fig:intro}
\end{figure}

Video anomaly detection (VAD) is an important public safety application, which aims to accurately detect abnormal events in videos in time. Since anomalies are too scattered and rare to be sufficiently and comprehensively collected, the setting of learning from abundant normal data is more practical, necessitating the semi-supervised approach~\cite{chandola2009anomaly, ramachandra2020survey}.

%Video surveillance systems are widely utilized in public and private areas for the safety of public life and assets. As a critical application, video anomaly detection (VAD) aims to accurately detect abnormal events in videos in time. 

Under this semi-supervised framework, reconstruction-based~\cite{gong2019memorizing, nguyen2019anomaly, zaheer2020old, hasan2016learning, cao2024context, wang2022video, ionescu2019object} and prediction-based~\cite{cao2023new, cao2024scene, liu2018future, cai2021appearance, liu2021hybrid, li2021variational, rodrigues2020multi, georgescu2021anomaly, park2020learning,  ristea2024self, lv2021learning} approaches are two of the most representative paradigms. They are based on the premise that anomalies are inherently harder to predict or reconstruct when compared to normal data~\cite{hasan2016learning, liu2018future}. However, the underlying assumption of these paradigms can not avoid undesired generalizations on anomalies because it only considers low-level vision information in videos. 

%Under this semi-supervised framework, the long-standing goal of most studies is to train a one-class classifier that faithfully learns normal data distribution while labeling events deviating from the learned distribution as abnormal. Reconstruction-based~\cite{gong2019memorizing, nguyen2019anomaly, zaheer2020old, hasan2016learning, cao2024context, wang2022video, ionescu2019object} and prediction-based~\cite{cao2023new, cao2024scene, liu2018future, cai2021appearance, liu2021hybrid, li2021variational, rodrigues2020multi, georgescu2021anomaly, park2020learning,  ristea2024self, lv2021learning} approaches are two of the most representative paradigms. They are based on the premise that out-of-distribution anomalies are inherently harder to predict or reconstruct when compared to in-distribution normal data~\cite{hasan2016learning, liu2018future}. However, the underlying assumption of these paradigms can not guarantee the avoidance of undesired generalizations on anomalies because it only considers low-level pixel details in videos. 

Recent years have witnessed the great progress of generative models~\cite{ho2020denoising, song2019generative, song2020score}. The powerful pattern coverage capability of generative models lays a new solid foundation for the VAD task: an out-of-distribution anomaly can be detected by its low likelihood under the statistical model of normal data. However, these likelihood-based methods build generative models for boosting visual pretext tasks~\cite{flaborea2023multimodal, yan2023feature} or assisting density estimation models~\cite{hirschorn2023normalizing, mahmood2020multiscale, micorek2024mulde}. Directly using a likelihood-based indicator for VAD remains challenging because it has some intrinsic limitations. We consider the Stein score~\cite{liu2016kernelized} here, which represents the gradient of log density. As shown in \cref{fig:intro}, the Stein score tends to be small at maxima, minima, and saddle points, which makes it difficult to detect anomalies that reside in local modes~\cite{mahmood2020multiscale}. As a result, solely relying on the likelihood is insufficient to make a distinction. In the task of VAD, these local models can be attributed to three gaps between likelihood and vision: scene gap regarding scene details~\cite{cao2023new, cao2024scene}, motion gap regarding motion details, and appearance gap regarding visual details.

%abnormal events may reside in local modes near the learned distribution due to their commonalities. As a result, ``flat" regions emerge from these local modes, where solely relying on the likelihood through distribution approximation is insufficient to make a distinction~\cite{mahmood2020multiscale}. In the task of VAD, these local models can be attributed to three gaps: scene gap regarding scene-dependent anomalies~\cite{cao2023new, cao2024scene}, motion gap regarding unusual movements, and appearance gap regarding rare appearances.

To fill these gaps, in this paper, we propose autoregressive denoising score matching (ADSM), which contains a novel model, a novel score function, and a novel autoregressive mechanism to take both the efficiency of likelihood estimation and the characteristic of video anomalies into account. Consider an input video sequence perturbed with various levels of Gaussian noises. We first integrate the denoising score matching into transformer-based architecture~\cite{vaswani2017attention} to construct a noise-conditioned score transformer (NCST) for approximating the Stein score function of noisy data. Then, we embed the scene information of the input sequence into our NCST to estimate a scene-dependent score. Next, we assign motion weights to the score function based on the difference between the first and last key frames~\cite{yang2023video} of the input sequence, guiding our method to focus on the motion consistency inherent in videos. The scene gap and motion gap are filled with a scene-dependent and motion-aware score function.

%Consider an input video sequence, we first perturb the input with various levels of Gaussian noises. Then, we integrate the denoising score matching into transformer-based architecture~\cite{vaswani2017attention} to construct a noise-conditioned score transformer (NCST) for approximating the Stein score function of noisy data. 

%Next, we embed the scene information of the input sequence into our NCST to estimate a scene-dependent score, filling the scene gap~\cite{cao2023new, cao2024scene, sun2023hierarchical} via modeling the relationships between video events and scenes. Finally, for the motion gap of anomalies, we assign motion weights to the score function based on the difference between the first and last key frames~\cite{yang2023video} of the input sequence, guiding our method to focus on the motion consistency inherent in videos.

Our main goal is to decrease the probability likelihood of these local modes while maintaining a high probability likelihood for normal data. Therefore, integrating the unaffected visual information to fill the appearance gap is effective. Finally, a novel autoregressive denoising score matching mechanism is developed for inference, in which we autoregressively inject intensifying Gaussian noise into the denoised data and estimate the corresponding score function. Along with each denoising process, we compare the denoised data with the original data to get a difference and aggregate it with the score function. In this way, not only is the abnormal context accumulated through the autoregressive mechanism, but the local mode is also suppressed by the visual information via an enhanced appearance perception. With all three gaps considered, we can compute a more suitable and comprehensive anomaly indicator.

%for a range of noise levels
% In particular, we develop a novel autoregressive denoising score matching mechanism for inference, in which we autoregressive inject intensifying Gaussian noise into the denoised data and estimate the corresponding score function. Unlike the core paradigm of a recent class of score-based generative models~\cite{micorek2024mulde, mahmood2020multiscale, song2019generative} that estimate the score function from the original noisy data, this autoregressive mechanism helps accumulate abnormal contexts of anomalies, resulting in a more compact distribution. Last but not least, we compare the denoised data with the original data to get a difference and aggregate it with the score function, compensating for the appearance gap via enhancing the perception of low-level pixel details. 

Our ADSM is the first score-based transformer for VAD. We modify the objective function to combine the denoising score matching with the characteristics of video anomalies for denoising diffusion transformers (DiTs)~\cite{peebles2023scalable}. We are inspired to utilize the DiTs for their improved mode coverage capabilities shown in a series of generative tasks~\cite{bao2023all, ma2024latte, mo2023dit}. Extensive experiments on Avenue~\cite{lu2013abnormal}, ShanghaiTech~\cite{luo2017revisit}, and NWPU Campus~\cite{cao2023new} datasets illustrate the superiority of our solution over other methods. To summarize, the main contributions of this paper are as follows:

%and the state-of-the-art performance
%such as image synthesis~\cite{bao2023all}, video synthesis~\cite{ma2024latte}, and 3D generation~\cite{mo2023dit}. 

\begin{itemize}
    \item A novel autoregressive denoising score matching mechanism for VAD, which autoregressively approximates the score function of noisy data and combines visual details based on the corresponding denoised data, building a good video anomaly detector.
    \item A novel transformer-based model for denoising score matching, which is the first score-based transformer for VAD, exploiting the enhanced mode coverage capabilities of diffusion probabilistic transformers.
    \item Two useful adjustment strategies making the score function scene-dependent and motion-aware, polishing a more suitable and comprehensive likelihood-based indicator for a range of noise levels.
    \item A thorough validation on Avenue, ShanghaiTech, and NWPU Campus datasets, where our method achieves the state-of-the-art performance in all three datasets, fully demonstrating the effectiveness.
\end{itemize}

%\item Three useful adjustment strategies settling the scene gap, motion gap, and appearance gap uniquely existing in VAD, separating the local mode of anomalies near the learned distribution for a better performance.
%which is based on autoregressively approximating the score function of noisy data to get a good video anomaly detector. which penetrates the high-level likelihood information and low-level pixel details to build a comprehensive likelihood-based indicator

\section{Related Work}
\label{sec:related}

\begin{figure*}
  \centering
  \includegraphics[width=1.9\columnwidth]{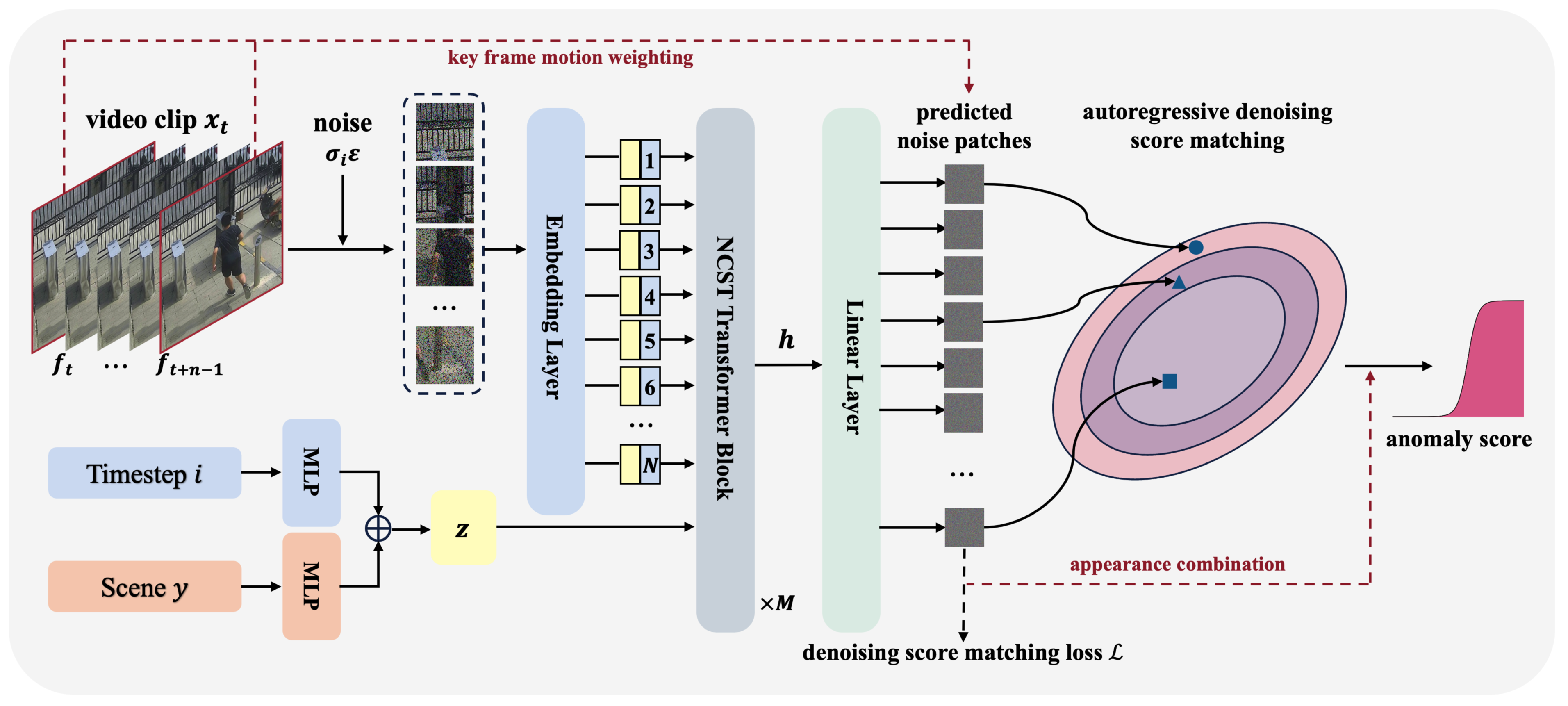}
  \caption{An overview of the proposed method. We implement autoregressive denoising score matching via a novel noise-conditioned score transformer (NCST), which consists of MLP, embedding layer, NCST blocks, and linear layer. When a video sequence is given, we first perturb it with noises of various levels and tokenize the noisy data into patches. Then, we feed them into the embedding layer to get patch embeddings. Next, the proposed NCST takes the patch embeddings (with position) and the overall condition \(z\) (generated based on diffusion time step \(i\) and scene class label \(y\)) as input to predict noise patches with the linear layer. Finally, we get anomaly scores through autoregressive denoising score matching.}
  \label{fig:pipeline}
\end{figure*}

%-------------------------------------------------------------------------
\subsection{Video Anomaly Detection Methods}
Prevalent VAD methods can be grouped into the semi-supervised category, weakly-supervised category~\cite{zhong2019graph, zhu2019motion}, and fully supervised category~\cite{zhou2016spatial}. We focus on the semi-supervised methods in this paper, which mainly contains distance-based~\cite{lu2022learnable, chang2020clustering, wu2019deep}, reconstruction-based~\cite{gong2019memorizing, nguyen2019anomaly, zaheer2020old, hasan2016learning, cao2024context, wang2022video, ionescu2019object}, and prediction-based~\cite{cao2023new, cao2024scene, liu2018future, cai2021appearance, liu2021hybrid, li2021variational, rodrigues2020multi, georgescu2021anomaly, park2020learning,  ristea2024self, lv2021learning, fang2020anomaly, luo2021future, song2019learning} methods. They train deep networks through pretext tasks, such as auto-encoding frames for reconstruction~\cite{gong2019memorizing, nguyen2019anomaly, zaheer2020old, hasan2016learning, ionescu2019object}, predicting future frames~\cite{cao2023new,  liu2018future, cai2021appearance,  rodrigues2020multi, park2020learning, lv2021learning, fang2020anomaly, luo2021future, song2019learning}, predicting future features~\cite{cao2024scene, liu2021hybrid, li2021variational}, learning a dictionary~\cite{lu2022learnable, chang2020clustering, wu2019deep}, and solving masked jigsaw puzzles~\cite{wang2022video, ristea2024self, georgescu2021anomaly}. Recently, the powerful mode coverage capability of the generative model enables a novel likelihood-based approach~\cite{hirschorn2023normalizing, micorek2024mulde, flaborea2023multimodal, yan2023feature} for VAD. They build generative models to learn the distribution of normal data for boosting visual pretext tasks~\cite{flaborea2023multimodal, yan2023feature} or assisting density estimation models~\cite{hirschorn2023normalizing, micorek2024mulde}.

Despite the impressive performance, directly using a likelihood-based indicator is suffering from the ``unseen" anomalies lying in the local mode. Our method is tailored to these regions through an autoregressive mechanism. 

% based on the characteristics of video anomalies
%use flow model~\cite{hirschorn2023normalizing}, diffusion model~\cite{flaborea2023multimodal, yan2023feature}, or denoising score matching model~\cite{micorek2024mulde}, to learn the distribution of normal data, by which the events with low likelihoods are detected as anomalies.

%Moreover, our method improves the mechanism of likelihood learning to form a more compact distribution in the VAD task.

%-------------------------------------------------------------------------
\subsection{Score-Based Generative Model}
Generative models have made great advances in recent years, in which diffusion models~\cite{ho2020denoising} integrated with Langevin dynamics for score matching ~\cite{song2019generative, song2020score} demonstrate remarkable advantages. The score function is defined as the gradient of the log density with respect to the input, which can be seen as a vector field pointing in the direction where the log data density grows the most. The good properties of the score function naturally serve the task of anomaly detection~\cite{rodrigues2020multi, cao2024scene, micorek2024mulde}. Cao \textit{et al}.~\cite{cao2024scene} use the score-based diffusion model to build a scene-conditioned auto-encoder in latent space for scene-dependent anomaly detection and anticipation. Rodrigues \textit{et al}.~\cite{rodrigues2020multi} approximate the score function for a range of noise levels for capturing multiscale gradient signals. Micorek \textit{et al}.~\cite{micorek2024mulde} find the log density performs better with video features at the cost of sacrificing the nonlinearity of the model. However, these methods suffer form the local modes of the score function, unable to make a distinction solely with the likelihood.

We dive into the denoising score matching mechanism and propose the autoregressive denoising score matching in VAD, which helps accumulate abnormal contexts and combine vision and likelihood information for complementarity.

%They use the norm of the log-density gradient as an anomaly indicator.
% They embed the scene condition into the score-based diffusion model for scene-dependent anomaly detection and anticipation. 
%An autoregressive score-matcher is more suitable for the task, as it helps accumulate abnormal contexts of anomalies, resulting in a more compact distribution.

%-------------------------------------------------------------------------
\subsection{Transformers in Diffusion Generation}
Diffusion transformers ~\cite{peebles2023scalable, bao2023all, bao2023one} have recently exhibited remarkable capability in generative tasks. Notably, the diffusion transformer (DiT)~\cite{peebles2023scalable} introduces a straightforward architecture for learning the denoising diffusion process in the latent space of Stable Diffusion~\cite{rombach2022high}. U-DiT~\cite{bao2023all} leverages a Vision Transformer (ViT)~\cite{dosovitskiy2020image} framework with extensive skip connections between shallow and deep layers for denoising the noisy image patches. UniDiffuser~\cite{bao2023one} devises a unified transformer tailored to diffusion models, enabling the simultaneous learning of all distributions to handle diverse input modalities effectively.

Motivated by the promising performance of diffusion transformers, we develop a novel diffusion transformer based on the denoising score matching for VAD. Without the complex generation process, we show that the score output itself estimated by the powerful DiT is a good video anomaly detector.

\section{Proposed Method}
\label{sec:method}

\cref{fig:pipeline} presents an overview of the proposed method. We perform anomaly detection directly on the video clip since the extracted features with diverse distribution from pretrained models highly account for the unstable performance of VAD methods. The space of the original video data enables us to focus on the characteristics of video anomalies.

%-------------------------------------------------------------------------
\subsection{Denoising Score Matching}

Score matching~\cite{hyvarinen2005estimation} is introduced to learn non-normalized statistical models, where the score is defined as the gradient of the log density \(\nabla_{x} log p(x)\) with respect to the data \(x\). Following work~\cite{vincent2011connection}, we can perturb the data \(x\) with \textit{iid} Gaussian noise and use score matching to directly train a score network \(s_{\theta}(\tilde{x})\) for estimating the score of the perturbed data distribution \(q_{\sigma}(\tilde{x})=\int q_{\sigma}(\tilde{x}|x) p(x)\), where \(\sigma\) is the noise level. The objective function is proved equivalent to matching the score of a non-parametric Parzen density estimator of the data:

\begin{equation}
  \frac{1}{2} \mathbb{E}_{q_{\sigma}(\tilde{x}|x)p(x)} [ || s_{\theta}(\tilde{x})-\nabla_{\tilde{x}} log q_{\sigma}(\tilde{x}|x) ||_2^2 ]
  \label{eq:dae}
\end{equation}

This approximation can be further used for generative models~\cite{song2019generative, song2020score}, which perturb data with a sequence of intensifying Gaussian noise and jointly estimate the score functions for all noisy data distributions by training a score network \(s_{\theta}(x,\sigma)\) conditioned on noise levels \(\sigma_i\) (\textit{i.e.,} \(\forall \sigma \in \{\sigma_i\}_{i=1}^L : s_{\theta}(x,\sigma)\approx\nabla_x log q_{\sigma}(x)\)). If we choose the noise distribution to be \(\mathcal{N}(\tilde{x}|x,\sigma^2 I)\), the score function is equivalent to \(\nabla_{\tilde{x}} log q_{\sigma}(\tilde{x}|x)=-\frac{\tilde{x}-x}{\sigma^2}\), which can be seen as a vector field that points in the direction of denoising~\cite{vincent2011connection, song2019generative}. Thus the objective function can be written as:

\begin{equation}
  \frac{1}{L} \sum_{i=1}^{L} \lambda(\sigma_i) [ \frac{1}{2} \mathbb{E}_{q_{\sigma_i}(\tilde{x}|x)p(x)} [ || s_{\theta}(\tilde{x},\sigma_i) + (\frac{\tilde{x}-x}{\sigma_i^2}) ||_2^2 ]]
  \label{eq:dsm}
\end{equation}
where \(\lambda(\sigma_i) > 0\) is a coefficient function and we set \(\lambda(\sigma_i)=\sigma_i^2\) following work ~\cite{song2019generative}. In the following subsections, we build a novel model, a novel score function, and a novel autoregressive mechanism to show that the score itself is a good video anomaly detector rather than utilizing it for generative models.

%we build a noise-conditioned score transformer for denoising score matching and combine it with the characteristics of anomalies in videos. Moreover, we propose a novel autoregressive denoising score matching mechanism to autoregressively estimate the score function for inference. We show that the score itself is a good video anomaly detector rather than utilizing it for generative models.

%-------------------------------------------------------------------------
\subsection{Noise-Conditioned Score Transformer}
In order to implement denoising score matching in the task of VAD, we need a model that can not only coordinate the complex spatiotemporal information of videos but also consider various conditions like noise levels. Attention-based architectures~\cite{vaswani2017attention} present an intuitive option for capturing long-range contextual relationships in videos. Moreover, DiT~\cite{peebles2023scalable} has demonstrated the possibility of incorporating the denoising process into the attention mechanisms. Therefore, we modify the denoising score matching formulation to train a novel noise-conditioned score transformer.

As shown in \cref{fig:pipeline}, let \(x_{t}=\{f_{t},\cdots,f_{t+n-1}\}\) be an input sequence of \(n\) frames at the time step \(t\), \(\{\sigma_i\}_{i=1}^{L}\) be a geometric sequence that satisfies \(\frac{\sigma_L}{\sigma_{L-1}}=\cdots = \frac{\sigma_2}{\sigma_1} > 1\). At noise level \(\sigma_i\), we perturb the input video sequence with Gaussian noise \(\epsilon\) to get \(\tilde{x}_{t}\), the distribution of which is \(q_{\sigma_i}(\tilde{x}|x)=\mathcal{N}(\tilde{x}|x,\sigma_i^2 I)\). We sample these \(n\) perturbed frames \(\tilde{x}_{t}\) and embed each individual frame into tokens using the uniform frame patch embedding method introduced in ViT~\cite{dosovitskiy2020image}. Let \(N\) be the number of non-overlapping patches of size \(d \times d \times c\) from each sequence \(\tilde{x}_{t}\), where \(c\) is the number of input channels and \(d\) is a hyperparameter that determines \(N\) with \(n\). Then, we get a set of tokens \(\{\tilde{P}_j\}_{j=1}^{N} \in \mathbb{R}^{d \times d \times c}\) in \(\tilde{x}_{t}\) and  \(\{P_j\}_{j=1}^{N} \in \mathbb{R}^{d \times d \times c}\) in \(x_{t}\). The target of our noise-conditioned score transformer \(s_\theta(P,\sigma)\) is to estimate the score functions for all noisy patches with the score matching objective in \cref{eq:dsm}. To that end, we change the objective to make a patch-wise denoising score matching function, in the sense of solving:

\begin{equation}
\begin{split}
  \mathcal{L}_{\text{NCST}}=&\frac{1}{L} \sum_{i=1}^{L}\{ \frac{\sigma_i^2}{2} \sum_{j=1}^{N} \mathbb{E}_{q_{\sigma_i}(\tilde{P}_j|x)p(x)} 
  \\&[ || s_{\theta}(\tilde{P}_j,\sigma_i) + (\frac{\tilde{P}_j-P_j}{\sigma_i^2}) ||_2^2 ] \}
  \label{eq:pdsm}
\end{split}
\end{equation}
Our noise-conditioned score transformer is implemented through the latent diffusion transformer (Latte)~\cite{ma2024latte}. We train the model conditioned on the diffusion time step \(i\) to control the noise level \(\sigma_i\) (embedded through an MLP). With attention-based architectures, the consecutive supervision of the multiscale patch-wise denoising process can effectively result in a conservative vector field, making the anomalies distinguishable.

\begin{figure}
  \centering
  \includegraphics[width=0.36\columnwidth]{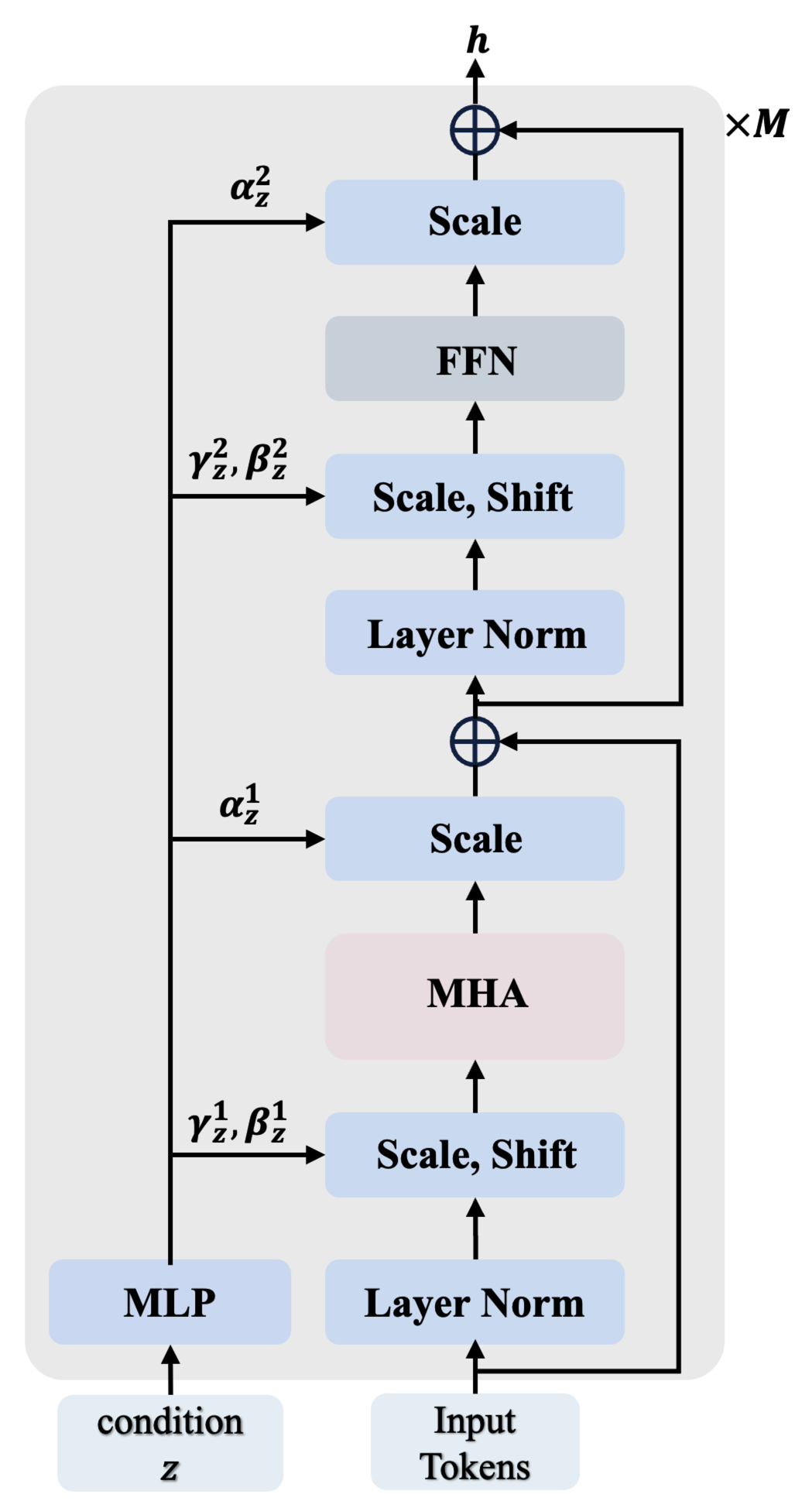}
  \caption{A diagram of our NCST block architecture used in \cref{fig:pipeline}. MLP, MHA, and FFP stand for the multi-layer perception layer, the feed-forward neural network, and the multi-head attention layer, respectively.}
  \label{fig:ncst}
\end{figure}

%-------------------------------------------------------------------------
\subsection{Scene-Conditioned Score Matching}
Scene-dependent anomalies~\cite{ramachandra2020survey} are catching increasing attention as a characteristic of anomalies in the field of VAD. Reasonably considering scene information of anomalies in videos can effectively enhance the performance of VAD methods~\cite{cao2023new, cao2024scene, sun2023hierarchical}. For each video sequence, we assign a scene class label \(y\) based on the scene  (\textit{i.e.}, the camera ID) to which it belongs. Then, we combine the scene class label \(y\) with the diffusion time step \(i\) to get the overall condition \(z\). We apply the scalable adaptive layer normalization (S-AdaLN) mechanism ~\cite{ma2024latte} to embed and transmit the overall condition \(z\), which is demonstrated to propagate the condition more effectively throughout the model. As shown in \cref{fig:ncst}, after getting the overall condition \(z\) based on the diffusion time step \(i\) and scene class label \(y\), we employ a MLP to compute \(\gamma_z^1\), \(\beta^1_z\), \(\alpha^1_z\), \(\gamma_z^2\), \(\beta^2_z\), and \(\alpha^2_z\) based on \(z\), resulting in a linear combination as follows:

\begin{equation}
\begin{split}
h' = h + \alpha_z^1 \text{MHA}[ \gamma_z^1 \text{LayerNorm}(h)+\beta_z^1 ]    
%\alpha_z h + \gamma_z \text{LayerNorm}(h)+\beta_z
%\text{S-AdaLN}(h,z)=\alpha_z h + \gamma_z \text{LayerNorm}(h)+\beta_z
\label{eq:sdaln}
\end{split}
\end{equation}
\begin{equation}
\begin{split}
h = h' + \alpha_z^2 \text{FFN}[ \gamma_z^2 \text{LayerNorm}(h')+\beta_z^2 ]  
%\alpha_z h + \gamma_z \text{LayerNorm}(h)+\beta_z
\label{eq:sdaln1}
\end{split}
\end{equation}
where \(h\) and \(h'\) are hidden outputs within the transformer blocks. MHA and FFP stand for the multi-layer perception layer ~\cite{vaswani2017attention} and the feed-forward neural network, respectively. With this more adaptive manner, we transmit the condition information to each transformer block, contributing to superior performance and faster model convergence.

%-------------------------------------------------------------------------
\subsection{Key Frame Motion Weighting}
Unlike the commonly used Fisher score \(\nabla_\theta log p_\theta(x)\) in statistics, the score considered here is a function of the input sequence \(x_t\) rather than the model parameter \(\theta\). Therefore, we can improve this score based on the characteristics of the input data to achieve better performance. Since the dynamic foreground and static background of video frames are not balanced~\cite{ramachandra2020survey} in the security scenes, the score tends to reflect more static information. To this end, additionally integrating the motion information into the score function is beneficial. Based on the video codec theory~\cite{liu2010key}, we propose to use the key frame difference between the first and last frame of the input sequence to assign weights to the score during denoising score matching.

% Especially in the scenes of security, the dynamic foreground and static background of video frames are not balanced~\cite{ramachandra2020survey}. As a result, the score tends to reflect more static information since it is a vector field that points to directions along which the probability density function has the largest growth rate. To this end, additionally integrating the motion information into the score function is better. We propose to use the key frame difference between the first and last frame of the input sequence to assign weights to the score during denoising score matching.

For every input sequence \(x_t\), we first compute the absolute difference between the key frames to get \(g_t = |f_{t+n-1}-f_t|\). Then, we divide it into non-overlapping patches \(\{G_j\}_{j=1}^N \in \mathbb{R}^{d \times d \times c}\). Next, we compute the maximum difference along the \(c\) dimension and take the average across channels for each patch, which can be written as:

\begin{equation}
  \upsilon_j = \frac{1}{c}\sum_{l=1}^{c} max ( G_{j,l} ), G_{j, l} \in \mathbb{R}^{d \times d}
  \label{eq:channel-wise}
\end{equation}
Finally, we normalize the value to compute a patch-wise weight and allocate it to our score function:

\begin{equation}
  \omega_j = \frac{\upsilon_j}{\sum_{j=1}^N \upsilon_j}
  \label{eq:weight}
\end{equation}
\begin{equation}
\begin{split}
    \mathcal{L}=&\frac{1}{L} \sum_{i=1}^{L} \{ \frac{\sigma_i^2}{2} \sum_{j=1}^{N} \mathbb{E}_{q_{\sigma_i}(\tilde{P}_j|x)p(x)} 
    \\&[ \omega_j || s_{\theta}(\tilde{P}_j,\sigma_i) + (\frac{\tilde{P}_j-P_j}{\sigma_i^2}) ||_2^2 ] \}
  \label{eq:weighted-score}
\end{split}
\end{equation}
Through implementing the motion-weighted denoising score matching process, the score output from our model is able to encapsulate more motion information.

%-------------------------------------------------------------------------
\subsection{Autoregressively Anomaly Scoring} \label{subsec:adsm}

\begin{figure}
  \centering
  \includegraphics[width=0.9\columnwidth]{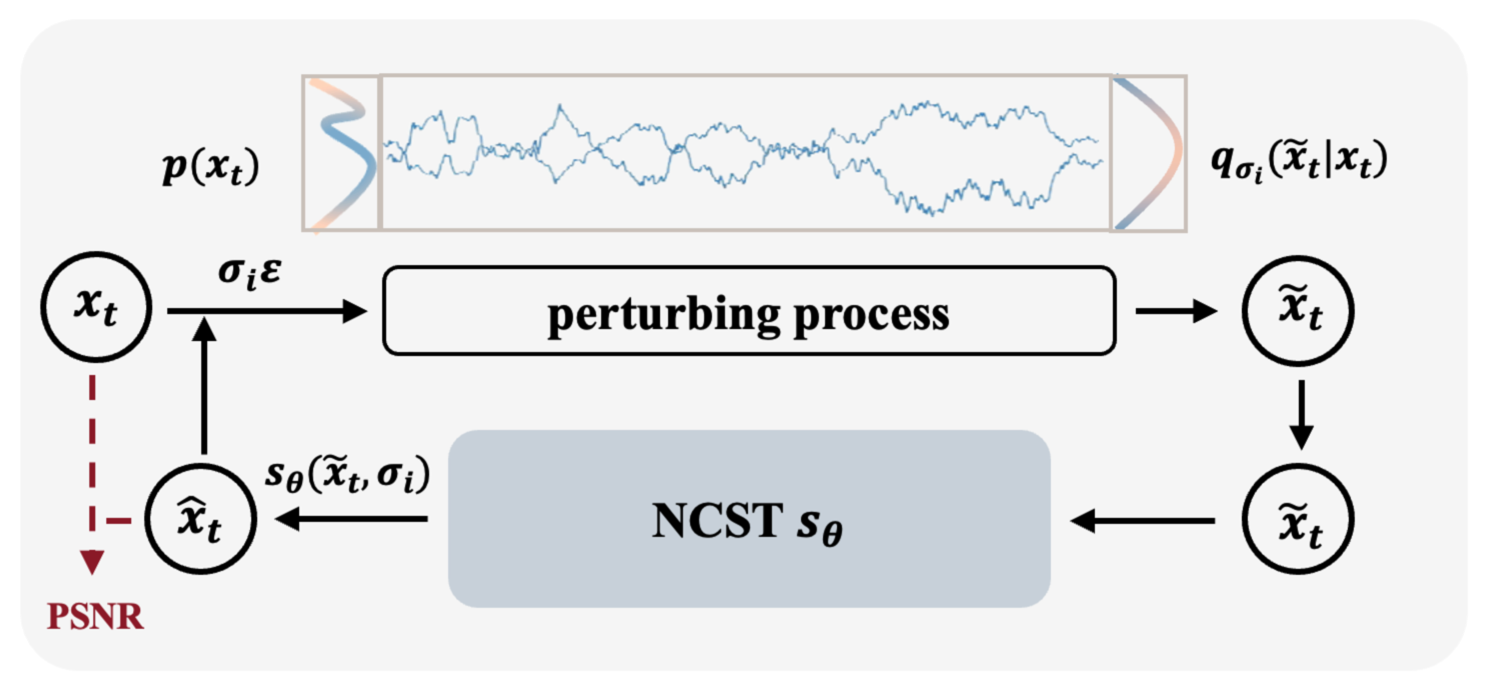}
  \caption{An overview of the proposed autoregressive denoising score matching process.}
  \label{fig:ar}
\end{figure}

For inference, we consider the L2-norm of the score: 

\begin{equation}
  ||s_{\theta}(x)||=||\nabla_x log p(x)||=||\frac{\nabla_x p(x)}{p(x)}||
  \label{eq:anomaly-score}
\end{equation}

In principle, the data density term in the denominator implies that normal events with high likelihoods yield a low norm, while anomalies with low likelihoods result in a high norm. However, the intrinsic limitations of the numerator mentioned in our motivation constrict the effectiveness of the score norm as a good anomaly indicator. To increase the norm of these local modes while maintaining a small norm for normal events, we propose an autoregressive denoising score matching mechanism for multiple noise levels. As shown in \cref{fig:ar}, for each perturbed \(\tilde{x}_{t}\) obtained by adding intensifying noise to the original data, the denoised data \(\hat{x}_t = \tilde{x}_t + \sigma_i^2 s_\theta(\tilde{x}_t, \sigma_i)\) is derived through jointly estimating the score function. We improve the mechanism by autoregressively applying intensifying noise to the denoised data instead of the original data and estimating the corresponding score function. This modification allows the abnormal context of anomalies to be accumulated, resulting in a more effective score function for VAD. 

Further, since traditional approaches based on unaffected visual pretext tasks have demonstrated fruitful progress, integrating visual information is prospective for complementarity. Therefore, in each denoising process, we utilize the PSNR value between \(x_t\) and \(\hat{x}_t\) as the denominator of the score norm. When the likelihood and appearance information in the numerator and denominator share consistency, the judgment of normal or abnormal events is highly probable and reliable. In contrast, when they diverge, the regions where the numerator misjudges are suppressed by the denominator, while the parts overlooked by the denominator are compensated by the numerator, resulting in a more robust and balanced score with the appearance gap filled. Finally, for every input sequence \(x_t\), we get \(L\) scores at the time step \(t\) from various noise level \(\sigma_i\):

%When the likelihood and appearance information located at the numerator and denominator share consistency, the judgment of normal or abnormal is highly probable and reasonable. When they go different ways, the area where the numerator misjudges will be suppressed by the denominator, and the parts where the denominator ignores can also be supplemented by the numerator.
%When the two values share consistency, the judgment of normal or abnormal is highly probable and reasonable. When the two values go different ways, the two types of information located at the denominator and numerator will complement each other. In this way, a more comprehensive score is obtained with an enhanced appearance perception, filling the appearance gap. 

\begin{equation}
  \text{score}_{i}(t)=\frac{||s_\theta(\tilde{x}_t, \sigma_i)||}{PSNR(\hat{x}_t, x_t)}
  \label{eq:psnr-score}
\end{equation}

During inference, we divide the video into \(\eta\) clips with \(T\) frames. Then, for each noise level \(\sigma_i\), we regard the maximum score as the score of the video clip to widen the discrimination between normal and abnormal events~\cite{yang2023video, cao2024scene}. Next, we normalize each \(\text{score}_i(t)\) to obtain anomaly scores \(S_{i}(t)\) in the range \([0,1]\) for quantifying the probability of anomalies occurring~\cite{mathieu2015deep}, which can be written as:

\begin{equation}
\begin{split}
S_{i}(t)&=\text{normalize}(\text{score}_{i}(t))
\\&=\frac{\text{score}_{i}(t)-\min\limits_{1\leq t\leq \eta T} \text{score}_{i}(t)}{\max\limits_{1\leq t\leq \eta T} \text{score}_{i}(t)-\min\limits_{1\leq t\leq \eta T} \text{score}_{i}(t)}
\end{split}
\label{eq:normalize}
\end{equation}
we weigh and average these anomaly scores \(S_{i}(t)\) to obtain an ultimate anomaly indicator, which is strong enough for VAD. The whole procedure is outlined in the Algorithm \ref{alg:infer}.

\begin{algorithm}[t]
    \SetAlgoLined %显示end
	\caption{Autoregressively anomaly scoring}%算法名字
	\KwIn{ the number of noise levels \(L\)\; 
               \(\epsilon \leftarrow \text{sample } \mathcal{N}(0,I)\)\;
               \(\sigma \leftarrow \text{sample log-uniform } (\sigma_1, \sigma_L)\)\;
               \(n\) input frames \(x_{t}=\{f_{t},\cdots,f_{t+n-1}\}\)\; 
               the score network \(s_{\theta}\)} %输入参数
	\KwOut{anomaly score \(S_{i}(t)\) \(\quad\quad\quad\quad\quad\quad\quad\quad\)} %输出
        \(x_{t}=\{f_{t},\cdots,f_{t+n-1}\}\)\;
        \(\dot{x_t}=x_{t}\)\;
        \For{\(i \in [1,L]\)}{
            \(\tilde{x}_{t} = \dot{x_t} + \sigma_i \epsilon\)\;
            \(\hat{x}_t=\tilde{x}_t+\sigma_i^2 s_\theta(\tilde{x}_t,\sigma_i)\)\;
            \(\text{score}_{i}(t)=||s_\theta(\tilde{x}_t, \sigma_i)|| / PSNR(\hat{x}_t, x_t)\)\;
            \(\dot{x_t}=\hat{x}_t\)\;
        }
        \(S_{i}(t)=\text{normalize}(\text{score}_{i}(t))\)\;
	return \(S_{i}(t)\)
    \label{alg:infer}
\end{algorithm}

\section{Experiment}
\label{sec:experiment}
%-------------------------------------------------------------------------
\begin{table}[t]
\caption{Comparison of different methods on the Avenue, ShanghaiTech, and NWPU Campus datasets in AUCs (\(\%\)) metric. The best result on each dataset is shown in bold. We also list the inputs of the methods, in which method without marking denotes single modal input. ``\(^\Diamond\)" and ``\(^+\)" represent multimodal inputs with optical flow and skeleton, respectively.}
\setlength\tabcolsep{1pt}
\centering
\begin{tabular}{ccccccc}
\hline
\multirow{2}{*}{Method} & \multicolumn{2}{c}{Avenue} & \multicolumn{2}{c}{ShanghaiTech} & \multicolumn{2}{c}{NWPU} \\ \cline{2-7} 
                        & Micro        & Macro       & Micro           & Macro          & Micro       & Macro      \\ \hline
\(^\Diamond\)FFP~\cite{liu2018future}                     & 84.9         & -           & 72.8            & -              & -           & -          \\
\(^\Diamond\)AMMC-Net~\cite{cai2021appearance}                & 86.6         & -           & 73.7            & -              & 64.5        & -          \\
\(^\Diamond\)HF\(^2\)-VAD~\cite{liu2021hybrid}                  & 91.1         & 93.5        & 76.2            & 83.1           & 63.7        & -          \\
\(^\Diamond\)VABD~\cite{li2021variational}                    & 86.6         & -           & 78.2            & -              & -           & -          \\ \hline
\(^+\)MTP~\cite{rodrigues2020multi}                     & 82.9         & -           & 76.0            & -              & -           & -          \\
\(^+\)STG-NF~\cite{hirschorn2023normalizing}                  & -            & -           & \textbf{85.9}            & -              & -           & -          \\
\(^+\)MoCoDAD~\cite{flaborea2023multimodal}                 & 89.0         & -           & 77.9            & -              & -           & -          \\ \hline
MemAE~\cite{gong2019memorizing}                   & 84.9         & -           & 72.8            & -              & 61.9        & -          \\
CAE~\cite{ionescu2019object}                     & 87.4         & 90.4        & 78.7            & 84.9           & -           & -          \\
MNAD~\cite{park2020learning}                     & 88.5         & -           & 70.5            & -              & 62.5        & -          \\
OG-Net~\cite{zaheer2020old}                  & -            & -           & -               & -              & 62.5        & -          \\
MPN~\cite{lv2021learning}                    & 89.5         & -           & 73.8            & -              & 64.4        & -          \\
SSMTL~\cite{georgescu2021anomaly}                 & 91.5         & 91.9        & 82.4            & 89.3           & -           & -          \\
LLSH~\cite{lu2022learnable}                    & 87.4         & 88.6        & 77.6            & 85.9           & 62.2        & -          \\
FBAE~\cite{cao2023new}                    & 86.8         & -           & 79.2            & -              & 68.2        & -          \\
FPDM~\cite{yan2023feature}                    & 90.1         & -           & 78.6            & -              & -        & -          \\
Two-stream~\cite{cao2024context}              & 90.8         & 93.0        & 83.7            & 90.8           & -           & -          \\
SDMAE~\cite{ristea2024self}                   & 91.3         & 90.9        & 79.1            & 84.7           & -           & -          \\
SSAE~\cite{cao2024scene}                    & 90.2         & 91.5        & 80.5            & 88.9           & 75.6        & 76.9       \\ \hline
ADSM (ours)             & \textbf{91.6}         & \textbf{94.2}        & 84.5            & \textbf{93.2}           & \textbf{76.9}        & \textbf{78.1}       \\ \hline
\end{tabular}
\label{tab:performance}
\end{table}

%-------------------------------------------------------------------------
\subsection{Experimental Setup} \label{subsec:set}
\textbf{Datasets.} We evaluate the performance of our method on three public datasets widely used in the VAD community. (1) Avenue dataset~\cite{lu2013abnormal} consists of  16 training videos and 21 testing videos. The abnormal events include running, throwing bags, child skipping, \textit{etc}, on the sidewalk. (2) ShanghaiTech dataset~\cite{luo2017revisit} contains 330 training videos and 107 testing videos. The abnormal events include affray, robbery, fighting, \textit{etc}, in 13 different scenes. (3) NWPU Campus dataset~\cite{cao2023new} is the largest semi-supervised VAD dataset with 305 training videos and 242 testing videos. The abnormal events include jaywalking, u-turn, \textit{etc}, in 43 different scenes. Notably, it is the first VAD dataset that takes the scene-dependent anomaly into account.

\noindent\textbf{Evaluation Metric.} We use the area under the curve (AUC) of the receiver operating characteristic (ROC) to evaluate the performance of our method. There are two methods to aggregate the AUC-ROC value over multiple videos in the literature: the micro score and the macro score~\cite{acsintoae2022ubnormal, barbalau2023ssmtl++, georgescu2021background, reiss2022attribute}. For the macro score, the AUC-ROC value is computed separately for each video and then averaged across all videos in the test set. The micro score computes the AUC-ROC jointly for all frames of all test videos. We report the results in terms of both metrics.

\noindent\textbf{Implementation Details.} The input frames of our model are the regions of \(160 \times 160\) pixels centered on objects that are detected by the pretrained ByteTrack~\cite{zhang2022bytetrack} implemented by MMTracking~\cite{mmtrack}. We expect our model to not only focus on target information but also consider some scene information on its own. If a given frame contains multiple targets, we will process each target separately. Our proposed model \(s_\theta\) receives an input sequence with 8 frames at a time and tokenizes the input sequence into \(N=8\times 100\) patches \(\{P_j\}_{j=1}^{N} \in \mathbb{R}^{d\times d\times c}\), where the patch size \(d\) is 16. Each model is trained for 100 epochs. The batch size is set to 20 and Adamax~\cite{kingma2014adam} is chosen as the optimizer. The initial learning rate is set to 0.0001 and gradually decays following the scheme of cosine annealing. During training, we sample the noise scale used for each batch element from the log-uniform distribution on the interval \([\sigma_1, \sigma_L]\) = \([0.001, 1.0]\). For evaluation, we use \(L = 20\) evenly spaced noise levels between \([\sigma_1, \sigma_L]\). The scene class label embedding and the diffusion time embedding, both with dimensions of 768, are concatenated for model learning and inference. All the experiments are performed on four NVIDIA GeForce RTX 4090 GPUs with pytorch~\cite{paszke2019pytorch} framework.

%\(\text{MLP}(y) \in \mathbb{R}^{768}\)
%\(\text{MLP}(i) \in \mathbb{R}^{768}\)
%Jigsaw~\cite{wang2022video}                  & \textbf{92.2}         & 93.0        & 84.3            & 89.8           & -           & -          \\

%-------------------------------------------------------------------------

\subsection{VAD Performance Benchmarking}
The performance comparison between our method and other existing methods in terms of the micro and macro scores on the Avenue, ShanghaiTech, and NWPU Campus datasets is present in~\cref{tab:performance}. We prefer the macro score, especially in datasets with multiple scenes like ShanghaiTech and NWPU Campus datasets, which can not be comprehensively and fairly judged by the micro score due to the fuzzy boundaries of scenes. Our method outperforms the others on the three datasets in terms of the macro score, all of which contain over 10 classes of abnormal events. Especially in the ShanghaiTech dataset, we achieve an obvious improvement (up to \(2.4 \%\)), demonstrating the effectiveness of our method. In the recently released NWPU Campus dataset, we achieve the best performance on both metrics, verifying the scalability and stability of our method. Compared with the works that use multimodal inputs, we only train our model on raw video clips and achieve comparable or even better performance. Compared with the work using the same single type of input, we achieve the best performance on the three datasets in terms of both the micro and the macro scores. The superior performance proves that our proposed autoregressive denoising score matching is a good anomaly detector for the challenging and complex VAD task. Notably, since the score is computed from random noise, it inherently carries some uncertainty. However, the qualitative analysis in \cref{subsec:adsm} confirms the validity of the denoising score matching mechanism, and we have maximally controlled this uncertainty through conditional embedding and the integration of visual information. Furthermore, numerous testing shows that the results exhibit minimal fluctuation (less than 1\%), ensuring that the reported outcomes are stable and reliable.

%We find that the relatively low performance on the Avenue dataset in terms of the micro score is mainly due to the inaccurate object tracking of the tracking algorithm, which is caused by the low resolution of this relatively older dataset.

%-------------------------------------------------------------------------
\subsection{Study on Latent Space}

%These methods build various architectures to learn the denoising diffusion process from a pre-trained variational auto-encoder model. 
Generative models in latent space have been widely adopted to compress redundant information of the original data and improve computational efficiency to generate high-fidelity images and videos~\cite{ma2024latte, rombach2022high, bao2023all, bao2023one}. Unlike traditional generative tasks, the task of VAD is more sensitive to the distribution of input data, which means less redundant information than usual. Under this circumstance, the VAD performance may be limited by a reduced low-dimensional latent space that is unable to preserve essential information of inputs. Moreover, our motivation and experiments also demonstrate the importance of considering the characteristics of original video data. A good VAD performance may be attributed to a good feature extractor rather than a suitable likelihood estimation strategy, which forces us to reconsider the necessity and feasibility of implementing the denoising score matching in the latent space of the pre-trained model. 

Here, we propose an analysis that assesses the performance of our method in the latent space. We use the prevalent variational auto-encoder of Stable Diffusion ~\cite{rombach2022high} pretrained on the CoCo dataset~\cite{lin2014microsoft} to produce a latent representation of the uncorrupted input sequence and we perform our ADSM in this latent space with the NCST as described in \cref{fig:pipeline}. \cref{tab:latent} illustrates the performance in terms of the micro and macro scores on ShanghaiTech and NWPU Campus datasets. We notice that the latent version underperforms the original version, which demonstrates our suspicion that the inconsistent distribution between the features extracted by the pre-trained model and the VAD dataset exacerbates the unstable performance of the latent version. The representation alignment may relieve this problem~\cite{yu2024representation}. Directly implementing our ADSM on the original data space is more intuitive with the advancement of computing resources and model capabilities.

Since the performance of the VAD method is unstable due to the inconsistent distribution of features extracted by pretrained models. For two likelihood-based baselines related to our method, one of which is implemented in latent space~\cite{micorek2024mulde} and the other~\cite{mahmood2020multiscale} is applied to image anomaly detection. We reproduced them under our settings for a fair comparison on ShanghaiTech and NWPU Campus datasets. As shown in \cref{tab:mulde}, our method outperforms other related methods, demonstrating that applying denoising score matching in the original data space is more effective. The original data space enables us to take both the efficiency of likelihood estimation and the characteristic of video anomalies into account, supporting our motivation.

%Especially in the NWPU Campus dataset that contains scene-dependent anomalies, our method achieves an obvious improvement (over \(3.8 \%\)) over the other methods via additionally taking the scene condition into account, supporting our motivation.

\begin{table}[t]
\caption{AUCs (\(\%\)) of our model in latent space and original data space on ShanghaiTech and NWPU Campus datasets.}
\tabcolsep=0.02\linewidth
\centering
\begin{tabular}{ccccc}
\hline
\multirow{2}{*}{Method} & \multicolumn{2}{c}{ShanghaiTech}                                 & \multicolumn{2}{c}{NWPU}                                        \\ \cline{2-5} 
                        & Micro                          & Macro                          & Micro                          & Macro                          \\ \hline
L-ADSM                  & 77.6                           & 87.5                           & 68.1                           & 73.2                           \\
ADSM                    & \textbf{84.5} & \textbf{93.2} & \textbf{76.9} & \textbf{78.1} \\ \hline
\end{tabular}
\label{tab:latent}
\end{table}

\begin{table}[t]
\caption{AUCs (\(\%\)) of different likelihood-based methods in the original data space on ShanghaiTech and NWPU Campus datasets.}
\centering
\begin{tabular}{ccccc}
\hline
\multirow{2}{*}{Method} & \multicolumn{2}{c}{ShanghaiTech} & \multicolumn{2}{c}{NWPU} \\ \cline{2-5} 
                        & Micro          & Macro          & Micro       & Macro      \\ \hline
MSMA~\cite{mahmood2020multiscale}                    & 81.3           & 87.5           & 72.3        & 75.1       \\
MULDE~\cite{micorek2024mulde}                   & 82.6           & 90.4           & 73.1        & 74.5       \\
Ours                    & \textbf{84.5}           & \textbf{93.2}           & \textbf{76.9}        & \textbf{78.1}       \\ \hline
\end{tabular}
\label{tab:mulde}
\end{table}

%-------------------------------------------------------------------------
\subsection{Ablation Study and Analysis}

In \cref{tab:ablation}, we show the performance of our model variants on the ShanghaiTech dataset. We sequentially add denoising score matching (DSM), autoregressive denoising score matching (ADSM), scene condition embedding (Scene), key frame motion weighting (Motion), and visual difference aggregating (Appearance) to our noise-conditioned score transformer. As shown in \cref{tab:ablation}, the autoregressive mechanism benefits the denoising score matching for VAD, as there is an average improvement of \(4.1\%\) in terms of the micro and macro scores. Although the ShanghaiTech dataset does not especially consider the scene-dependent anomalies, there is still an improvement of \(2.2\%\) in terms of the macro score, which indicates that our model is equipped with an enhanced perception of scene information. Taking motion consistency into account also proves useful with an average improvement of \(2.9\%\), demonstrating that our method reasonably correlates the spatiotemporal information of videos for VAD. Aggregating the difference between the denoised data and the original data helps the most in computing a more distinguishable anomaly indicator based on the score function, resulting in a significant improvement of \(5.7\%\) (\(5.8\%\)) in terms of the micro (macro) score. The result demonstrates that aggregating unaffected visual information is complementary. There is an additional improvement of \(2.4\%\) (\(3.3\%\)) in terms of the micro (macro) score when we integrate these model variants, indicating that each variant of our model is complementary to each other.

%We analyze that this aggregation comprehensively considers the high-dimensional distribution information and the low-dimensional pixel detail of the video data. 
%, well coupling the relationship between scene information and anomalies.

\begin{table}[t]
\caption{AUCs(\(\%\)) of our model variants on the ShanghaiTech dataset.}
\centering
\tabcolsep=0.007\linewidth
\begin{tabular}{c|ccccc|cc}
\hline
\multirow{2}{*}{} & \multirow{2}{*}{DSM} & \multirow{2}{*}{ADSM} & \multirow{2}{*}{Scene} & \multirow{2}{*}{Motion} & \multirow{2}{*}{Appearance} & \multicolumn{2}{c}{ShanghaiTech} \\ \cline{7-8} 
                  &                      &                       &                       &                     &                     & Micro          & Macro          \\ \hline
1                 & \checkmark                    &                       &                       &                     &                     &    73.9            &   78.5             \\
2                 & \checkmark                    & \checkmark                     &                       &                     &                     &        76.4        &     84.1           \\
3                 & \checkmark                    & \checkmark                     &    \checkmark                   &                    &                     &       76.9         &    86.3            \\
4                 & \checkmark                    & \checkmark                     &                      &  \checkmark                   &                     &       78.6         &    87.7            \\
5                 & \checkmark                    & \checkmark                     &                       &                     & \checkmark                   &       82.1         &    89.9            \\
6                & \checkmark                    & \checkmark                     & \checkmark                     & \checkmark                   & \checkmark                   & \textbf{84.5}           & \textbf{93.2}           \\ \hline
\end{tabular}
\label{tab:ablation}
\end{table}

%-------------------------------------------------------------------------
\subsection{Running Speed and Model Size}

We perform all the experiments on four NVIDIA GeForce RTX 4090 GPUs with the pytorch framework. Our method is implemented purely on the raw video clip without an extra feature extractor. Furthermore, we utilize the powerful generative model only for a score rather than high-fidelity images or videos, which greatly simplifies the denoising process. This represents a very small computational cost. The running time of our ADSM is primarily determined by the size of our noise-conditioned score transformer. In our setup described in \cref{subsec:set}, our ADSM uses an NCST with 130M parameters and takes less than 20 milliseconds to process an input sequence of 8 frames, which enables video anomaly detection at 50 FPS. Even considering the additional consumption of the object detection model~\cite{zhang2022bytetrack, mmtrack}, our method can still maintain a speed around 35 FPS, suitable for a real-time monitoring system typically around 30 FPS. A more detailed description and comparison are presented in the supplementary material.

%In a real-time monitoring system typically around 30 FPS, our proposed method is completely suitable. A more detailed description and comparison are presented in the supplementary material.

%-------------------------------------------------------------------------
\subsection{Qualitative Results}
\begin{figure}
\centering
\includegraphics[width=0.95\columnwidth]{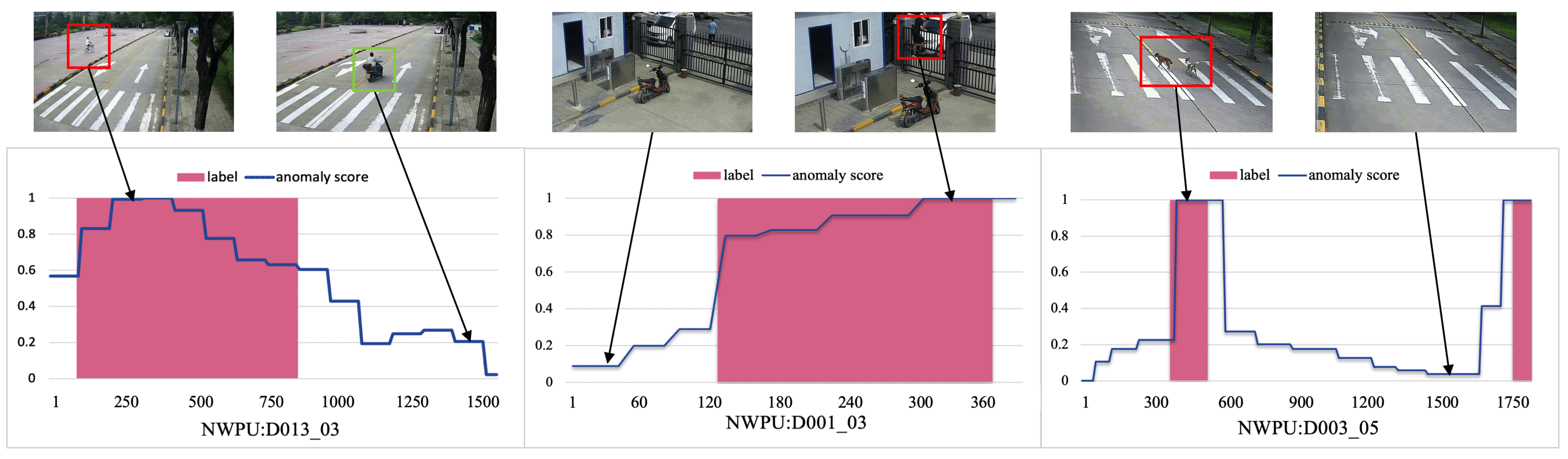}
\caption{A diagram of visualization on NWPU Campus dataset. Best viewed in color.}
\label{fig:visual}
\end{figure}

We visualize the score curves of our method on the NWPU Campus dataset for qualitative studies. We select three videos that contain scene-dependent anomalies (the cyclist on the square in the ``D013\_03"), motion anomalies (the climber in the ``D001\_03"), and appearance anomalies (the puppies in the ``D003\_05"), respectively. As shown in \cref{fig:visual}, concerning the scene-dependent cyclist anomaly that is abnormal on the square while normal on the road, our method makes a timely detection without false alarms. As for the abnormal motion of the climber and the abnormal appearance of the puppies, the anomaly score output by our method exhibits a sharp increase upon the occurrence of abnormal events and returns to flat when the abnormal events disappear. These results show that our method captures the characteristics of video anomalies in terms of scene, motion, and appearance. More visualization examples and analyses are provided in the supplementary material.

%The higher score represents the higher probability of anomaly.
%Video ``D013\_03" contains a scene-dependent anomaly, which is normal on the road while abnormal on the square.

% %-------------------------------------------------------------------------
% \subsection{Study on Density Estimation}

% aaaaaa
\section{Conclusion}
\label{sec:conclusion}

In this paper, we propose autoregressive denoising score matching (ADSM), which contains a novel model, a novel score function, and a novel autoregressive mechanism to not only utilize the efficiency of likelihood estimation but also take characteristics of video anomalies into account. We first design a novel noise-conditioned score transformer (NCST), which is the first score-based transformer for VAD, combining the efficiency of denoising score matching with the powerful diffusion probabilistic transformers. Next, we embed the scene condition and assign motion weights based on inputs to form a scene-dependent and motion-aware score function. Last but not least, we propose a novel mechanism to autoregressively approximate the score function and combine visual details based on the corresponding denoised data for anomaly accumulation and vision enhancement. Extensive experiments on three datasets verify the effectiveness of our method.

\section{Acknowledgments}
\label{sec:acknowledgment}

This work is supported by National Natural Science Foundation of China (No.62376217, 62301434), Young Elite Scientists Sponsorship Program by CAST (2023QNRC001), Key R\&D Project of Shaanxi Province (No. 2023-YBGY-240), and Young Talent Fund of Association for Science and Technology in Shaanxi, China (No. 20220117)

{
    \small
    \bibliographystyle{ieeenat_fullname}
    \bibliography{main}
}

% WARNING: do not forget to delete the supplementary pages from your submission 
%\input{sec/X_suppl}

\end{document}